\documentclass[10pt,twocolumn,letterpaper,pagenumbers]{article}

\usepackage{cvpr}              

\usepackage{graphicx}
\usepackage{amsmath}
\usepackage{amssymb}
\usepackage{booktabs}
\usepackage{multirow}

\usepackage[pagebackref,breaklinks,colorlinks]{hyperref}
\usepackage{color}

\def\cvprPaperID{1301} 
\def\confName{CVPR}
\def\confYear{2022}

\begin{document}

\title{One to Transfer All: A Universal Transfer Framework for Vision Foundation Model with Few Data}

\author{
Yujie Wang$^{1}$\thanks{Equal Contribution.} \quad Junqin Huang$^{1}$$^{*}$ \quad Mengya Gao$^{1}$$^{*}$ \quad Yichao Wu$^{1}$$^{*}$ \\ 
Zhenfei Yin$^{2}$ \quad Ding Liang$^{1}$ \quad Junjie Yan$^{1}$ \vspace{8pt}\\
$^{1}$SenseTime Research \qquad\qquad $^{2}$Shanghai AI Laboratory\qquad \\
\hspace{0.7in}{\tt\small \{wangyujie,huangjunqin,gaomengya,wuyichao,liangding,yanjunjie\}@sensetime.com}\\
{\tt\small yinzhenfei@pjlab.org.cn}\\
}

\maketitle

\begin{abstract}
The foundation model is not the last chapter of the model production pipeline.
Transferring with few data in a general way to thousands of downstream tasks is becoming a trend of the foundation model's application.
In this paper, we proposed a universal transfer framework: One to Transfer All (OTA) to transfer any Vision Foundation  Model (VFM) to any downstream tasks with few downstream data.
We first transfer a VFM to a task-specific model by Image Re-representation Fine-tuning (IRF) then distilling knowledge from a task-specific model to a deployed model with data produced by Downstream Image-Guided Generation (DIGG).
OTA has no dependency on upstream data, VFM, and downstream tasks when transferring.
It also provides a way for VFM researchers to release their upstream information for better transferring but not leaking data due to privacy requirements.
Massive experiments validate the effectiveness and superiority of our methods in few data setting.
Our code will be released.
\end{abstract}

\section{Introduction}
\label{sec:intro}
Large pre-trained model is turning into the foundation not only in Nature Language Processing (NLP)\cite{devlin2018bert,brown2020language} but also in Computer Vision (CV)\cite{kolesnikov2020big,he2020momentum}, a.k.a Foundation Models\cite{bommasani2021opportunities}.
Thousands of downstream task models can be transferred from these foundation models, and get fairly well performance with few data in a general way\cite{liu2021gpt,kolesnikov2020big}.
It seems that with a wonderful foundation model and few data, simply fine-tuning is sufficient for transferring.
However, it's not true for the vision foundation model (VFM).

\begin{figure}[h]
  \begin{center}
   \includegraphics[width=0.9\linewidth]{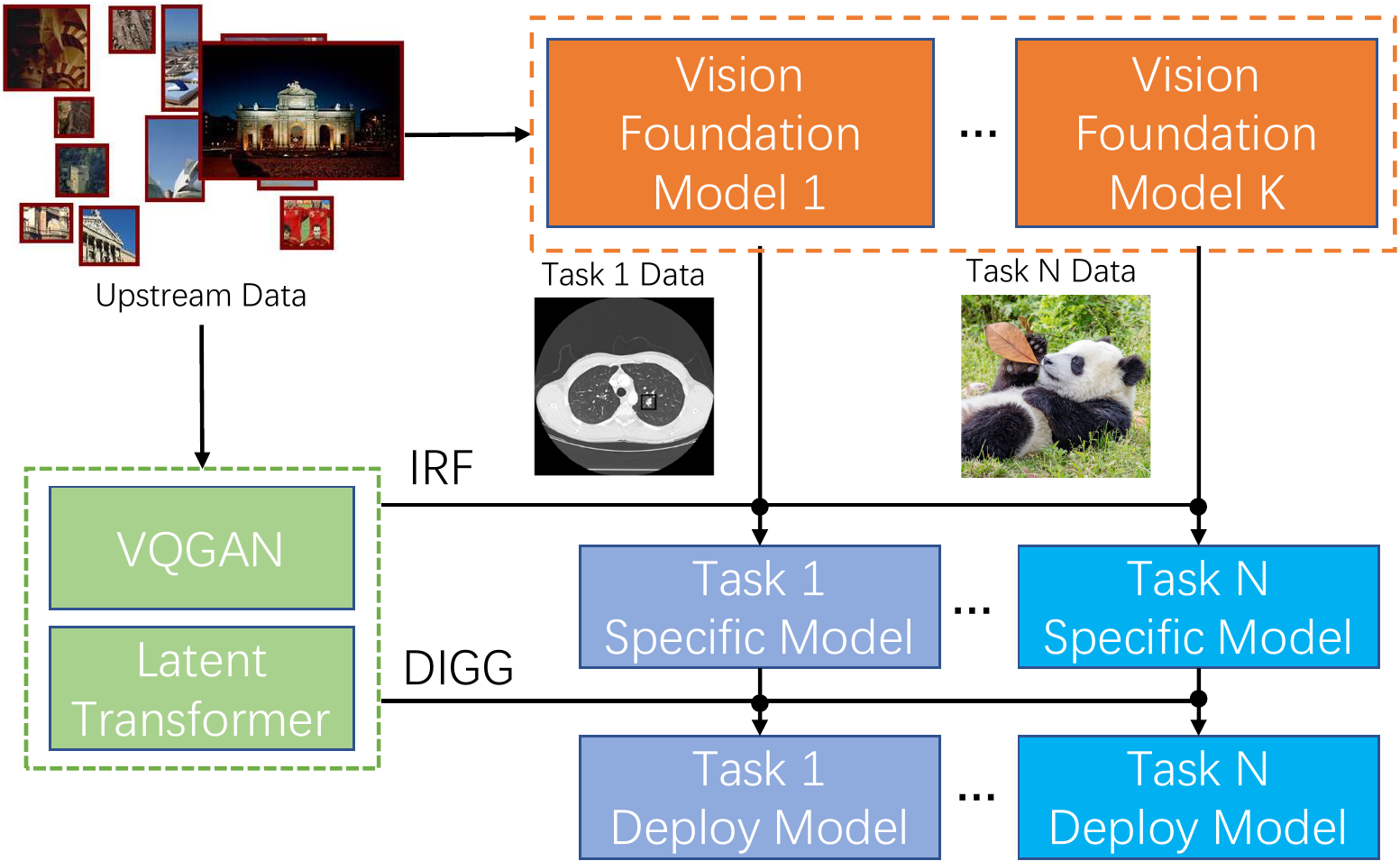}
    \end{center}
    \caption{Overview of our One to Transfer All (OTA) framework. The proposed Image Re-representation Fine-tuning (IRF) transfers a vision foundation model into a task-specific model, then the proposed Downstream Image-Guided Generation (DIGG) is used to generate images for distillation with few downstream data. With the proposed method, only one generation model (including VQGAN and latent transformer) is needed to transfer knowledge from any vision foundation model to any downstream task with few data. Best viewed in color.}
\label{fig:whole_pepeline}
\end{figure}

First, the performance of commonly used fine-tuning is not consistently higher than linear probe for downstream tasks\cite{ericsson2021well}, which indicates naive transfer methods are not satisfactory.
Especially with few data, it's hard to simultaneously adapt the VFM to fit downstream domain and optimize new parameters for downstream tasks like object detection’s head.
Second, strong transferability and high resource cost are two sides of the same coin for foundation models.
Smaller models are still indispensable not only for edge computing/AIoT devices but also for cloud computing to limit carbon emission.
So how to transfer knowledge from a large foundation model to a relatively small task-specific model is an inevitable challenge, not to mention with few downstream data and inaccessible upstream data for privacy requirements. 
Besides, a universal transfer method with minimal dependency on upstream data, VFMs, and tasks is also expected, which is also noticed in \cite{bommasani2021opportunities} for  adaptation evaluation.
To overcome such challenges, two vital issues should be considered.

The first issue is how to transfer the VFM into specific downstream tasks with few data in a universal way.
Many of recent top-performance transfer learning methods\cite{houlsby2019parameter,guo2019spottune,guo2020adafilter,ro2021autolr} indeed obtained prominent improvements.
However, these methods did not leverage the implicated information in the upstream training nor considered the insufficiency of downstream data for few data scenarios.

The second issue is how to transfer knowledge from a large model into a relatively small model with few data.
There are already many approaches proposed to transfer a big model into a small model in the literature, like pruning\cite{guo2020dmcp,enderich2021holistic} and knowledge distillation\cite{hinton2015distilling,tian2019contrastive}, but in the few data setting, these methods may be failed due to scarce and sparse data points.
Few existing works~\cite{Yin_2020_CVPR,rajasegaran2020self} try to relieve this issue.
But these methods are suitable for classification tasks and not trivial for other types of tasks such as object detection and semantic segmentation.

In this paper, we introduce a universal transfer framework, named One to Transfer All (OTA), for vision foundation models to overcome the two issues above. The framework firstly transfers a VFM to a task-specific model by our proposed Image Re-representation Fine-tuning (IRF), then transfers knowledge from this model to a deployed model with data produced by our proposed Downstream Image-Guided Generation (DIGG), as illustrated in Figure~\ref{fig:whole_pepeline}.
In OTA, a VQGAN (including encoder, decoder, and codebook) and latent transformer~\cite{esser2021taming} are trained with upstream data for transferring.
In the proposed IRF, we decouple the task-specific parameter's learning and domain adaptation.
Downstream images are first re-represented by VQGAN to pull the domain gap between the downstream data and upstream data, thus getting high-quality representation from a fixed VFM, which is used only to optimize task-specific parameters. Then original downstream images are used to optimize the whole model for adaptation.
As for DIGG, samples are generated by the latent transformer and VQGAN for knowledge distillation on condition of giving part of any downstream image.
The proposed framework has no dependency on the VFM's architecture, training strategy, or downstream tasks. With the trained VQGAN and latent transformer, any downstream task can be adapted by our methods, actually realizing \textit{one to transfer all}.
Exhaustive experiments prove the effectiveness of our methods.

Our contributions are summarized as follows:

\begin{itemize}
    \item We discussed problems in current model production pipelines with VFMs, and proposed a universal transfer framework to transfer knowledge from a VFM to a deployed model.
    \item We proposed a novel transfer method, named Image Re-representation Fine-tuning, to transfer knowledge from a vision foundation model to a task-specific model. It is simple, effective and has no dependency on upstream data, VFM and downstream tasks when transferring with few data.
    \item We proposed Downstream Image-Guided Generation method to relieve the lack of downstream data for distilling knowledge from large models.
    \item Our framework also provide an easy way for vision foundation model researchers to release their upstream information for better downstream transferring and not leaking their data.
\end{itemize}

\section{Related Work}

\textbf{Vision Foundation Model.}
The vision foundation model~\cite{bommasani2021opportunities} has a large family.
Pre-trained models in a large-scale dataset can be served as foundation models, no matter what learning paradigms or model architectures are adopted. 
Commonly used learning paradigms include supervised learning~\cite{he2016deep}, weakly supervised learning~\cite{mahajan2018exploring}, self-supervised learning~\cite{he2020momentum,chen2020simple}, and semi-supervised learning~\cite{sohn2020fixmatch}.
Currently, some technologies are borrowed from NLP, such as masked image modeling~\cite{bao2021beit}, auto-regressive modeling~\cite{pmlr-v119-chen20s}.
Besides, multi-modal foundation models also spring up, like 
CLIP~\cite{radford2021learning} and  WenLan~\cite{huo2021wenlan}.
In this paper, we did not focus on designing a transfer method for a specific vision foundation model but proposed a general method that can be adapted to any of these foundation models, especially in few data setting.

\textbf{Transfer Learning.}
Transfer learning aims at improving performance on target domains by transferring the knowledge from source domains or source foundation models. To achieve this, some of the recent top-performance transfer learning methods rely on the modification of the pre-trained model's architecture\cite{ramachandran2016unsupervised,houlsby2019parameter}, and some try to find important features and maintain them or only update part of weights more or less heuristically\cite{kirkpatrick2017overcoming,chronopoulou2019embarrassingly,howard2018universal,guo2019spottune,guo2020adafilter,ro2021autolr}.
All these methods indeed obtained prominent improvements, but their universality is not satisfactory.

Another term commonly used in the transfer learning area is domain adaptation.
Prior attempts in domain adaptation focus on finding an effective distribution matching strategy with the knowledge from the source domain and target domain, which can be defined as closed set domain adaptation\cite{long2015learning,courty2016optimal} or partial domain adaptation\cite{cao2018partial, zhang2018importance}. Compared with those approaches, our method has no dependency on source domain data when transferring. Works in \cite{kundu2020universal, feng2020kd3a, ahmed2021unsupervised, liang2020we} discussed the problem of source domain free adaptation which also assumes no prior knowledge on the source domain. However, these works do not give consideration to a unified transfer pipeline which is difficult to generalize to different types of downstream tasks.

\textbf{Distillation with Few Data.}
Knowledge Distillation (KD) is proposed by Hinton et al.~\cite{hinton2015distilling}.
In past years, it has been carried forward by many works\cite{peng2019correlation,tian2020contrastive}.
However, most of these works have a hypothesis that the data used for distillation is sufficient.
But the situation may change when we need to distill knowledge with few data.
Yin et al.\cite{Yin_2020_CVPR} proposed a method to generate images from the teacher without using real images, and the generated images are used to distill knowledge. Jathushan et al.\cite{rajasegaran2020self} proposed to exploit self-supervision for model learning.
These methods are suitable for classification tasks but are not trivial to transfer to other tasks.
In this paper, we followed the image generation path, by using our proposed DIGG method, which is more effective and efficient than previous methods.

\section{Method}

\begin{figure}
  \begin{center}
   \includegraphics[width=0.9\linewidth]{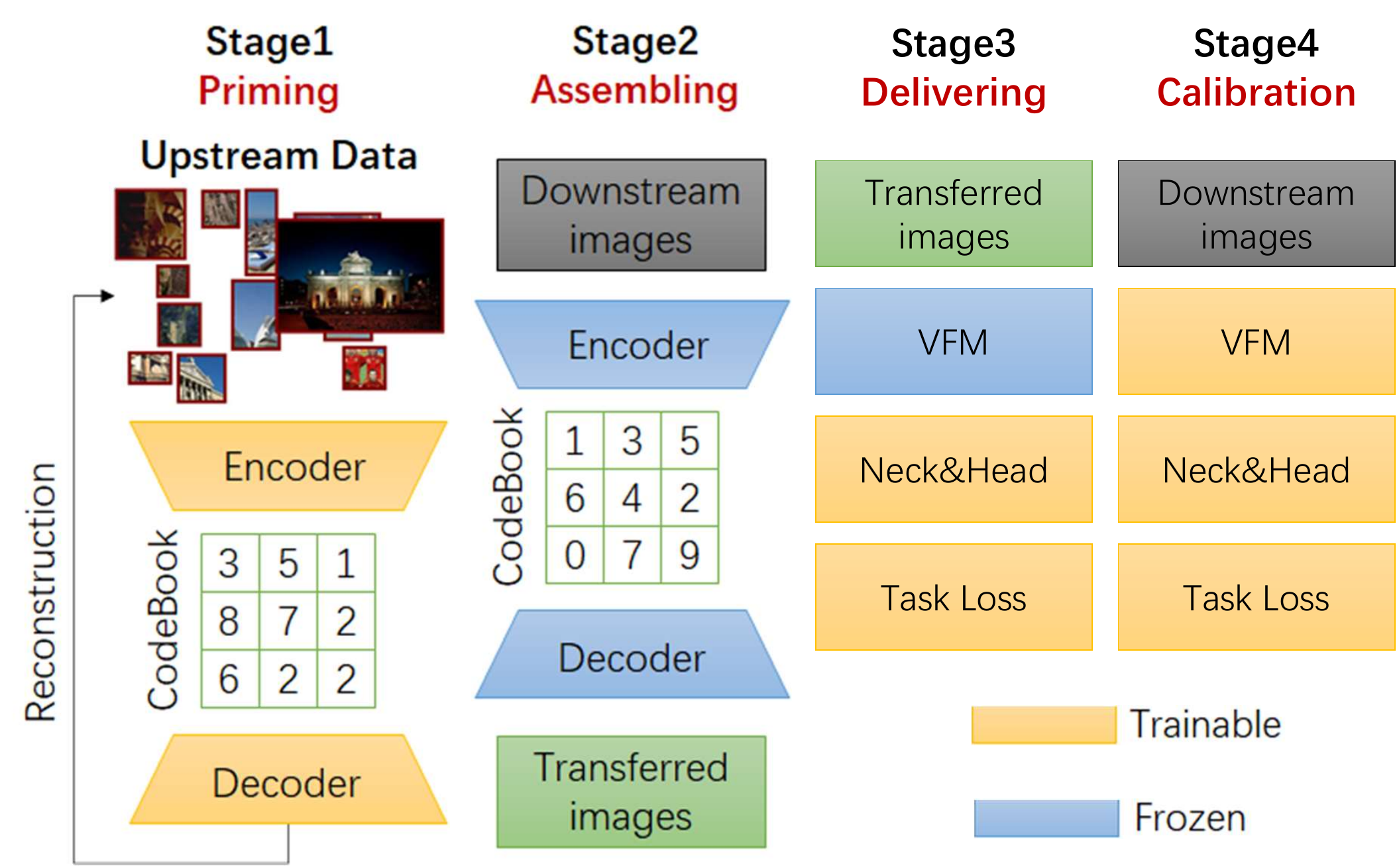}
    \end{center}
    \caption{Overview of IRF. A VQGAN model is firstly trained with upstream data in stage1. Then downstream data is re-represented by it in stage2. After that, only the new task-specific parameters are trained by re-represented images in stage3 and the whole model is fine-tuned by original downstream data in stage4.}
\label{fig:transfer-vp-pipeline}
\end{figure}

\subsection{Image Re-representation Fine-tuning}
\label{sec:irf}

When available downstream data is not sufficient, we find it's hard to jointly optimize new task-specific parameters and perform domain adaptation, e.g. performance of fine-tuning sometimes is inferior to linear probe at Table~\ref{tab:cls-downstream-transfer}.
Two issues may cause this phenomenon.
The first is that optimizing a large number of parameters by few data is too complex.
The second is the supervision collapse problem~\cite{doersch2020crosstransformers}.
Prediction errors will be backpropagated to a ``perfectly'' trained VFM and broke up the representation. There is not enough data and supervision signal to rebuild high-quality representation.
So we decouple the new task-specific parameters learning and domain adaptation, by firstly freezing the VFM and only optimizing new task-specific parameters.
However, the domain gap commonly exists between downstream data and upstream data, which makes downstream features extracted by the frozen VFM in low quality (may not precisely represent downstream data itself), leading to inferior optimization for new task-specific parameters.
With the fixed VFM and task loss, we intend to transfer downstream data to a new representation that has a small domain gap with upstream data, thus the feature of which is more reliable and informative.
Noticing that the high-level content of different images for different tasks may have large variance, we mainly focus on low-level transformation. The IRF has four stages as follows:

\textbf{Stage 1 Priming.} In this stage, we encode information of upstream data into a generative model by training a VQGAN~\cite{esser2021taming}, which will encode domain information of upstream data implicitly. The generative model is independent of the VFM, so its architecture is not needed to keep the same with the VFM.
Besides, it only needs to be trained once and can be reused by any downstream task.
So it both exploits upstream information and decouples the following stage with the VFM.

\textbf{Stage 2 Assembling.} In this stage, we will reconstruct downstream data by the generative model trained in the first stage, which can be seen as a re-representation process. The generative model is fixed in this stage because optimizing it with downstream data will lose the encoded upstream information.
We suppose a model can extract reliable features, i.e. a feature that can express the image's information precisely and sufficiently if an image is in a similar domain with its training data.
Then after the re-representation, features of transferred images extracted by the VFM will be more reliable for successive optimization.

\textbf{Stage 3 Delivering.}
Different downstream tasks need different task-specific layers to fulfill their jobs which are commonly not provided by the VFM, e.g. classifier for classification tasks, neck and head structure for detection tasks, etc.
How to optimize them into a proper state for further optimization is important especially in few data settings.
In this stage, we fix the VFM and only optimize the new task-specific layers with transferred images obtained in stage 2.
Using transferred images and fixing the VFM can keep the feature reliable after the extraction by the VFM, as a result, successive parts will get high-quality inputs and will be optimized into better local minima.

\textbf{Stage 4 Calibration.}
After the above three stages, the model is already in a not bad state, but it still lacks global optimization for domain adaptation of the VFM. 
And re-represented images may lose information in the reconstruction process that hinder performance boost further.
Besides, if we stopped at stage 3, when inference the model, the data should also be transferred, which is not proper for further usage.
So we calibrate the domain into the downstream domain thoroughly in this stage.
Original downstream images are used and all parameters are optimized in this stage.

\subsection{Downstream Image-Guided Generation}
\label{sec:digg}
One advantage of VFM is its great transferability with few data.
However, in few data setting, it is difficult to transfer knowledge from a large model to a small model by knowledge distillation as the student model can only mimic from a few data points, which will hinder its generalization performance.
A straightforward solution is to draw support from other academic datasets. But for different downstream tasks, using data that follows the downstream data distribution should get better results and it's hard to choose a suitable academic dataset for each downstream task.
In this paper, we in turn resort to conditioned image synthesis to alleviate this phenomenon.

As we mentioned before, a VQGAN\cite{esser2021taming} model is trained with upstream data, which consists of an encoder $\mathbf{E}$, a decoder $\mathbf{D}$, and a codebook $\mathbf{Z}$. 
Besides, a latent transformer~\cite{vaswani2017attention}, i.e. a next-index prediction model, is also trained with upstream data.
With the encoder $\mathbf{E}$, we can represent an image by a sequence of indices $s \in \{ 0,...,|Z|-1 \} $ from the codebook.
And the decoder $\mathbf{D}$ can easily map the codebook indices back to an image.

\begin{figure}
  \begin{center}
   \includegraphics[width=0.8\linewidth]{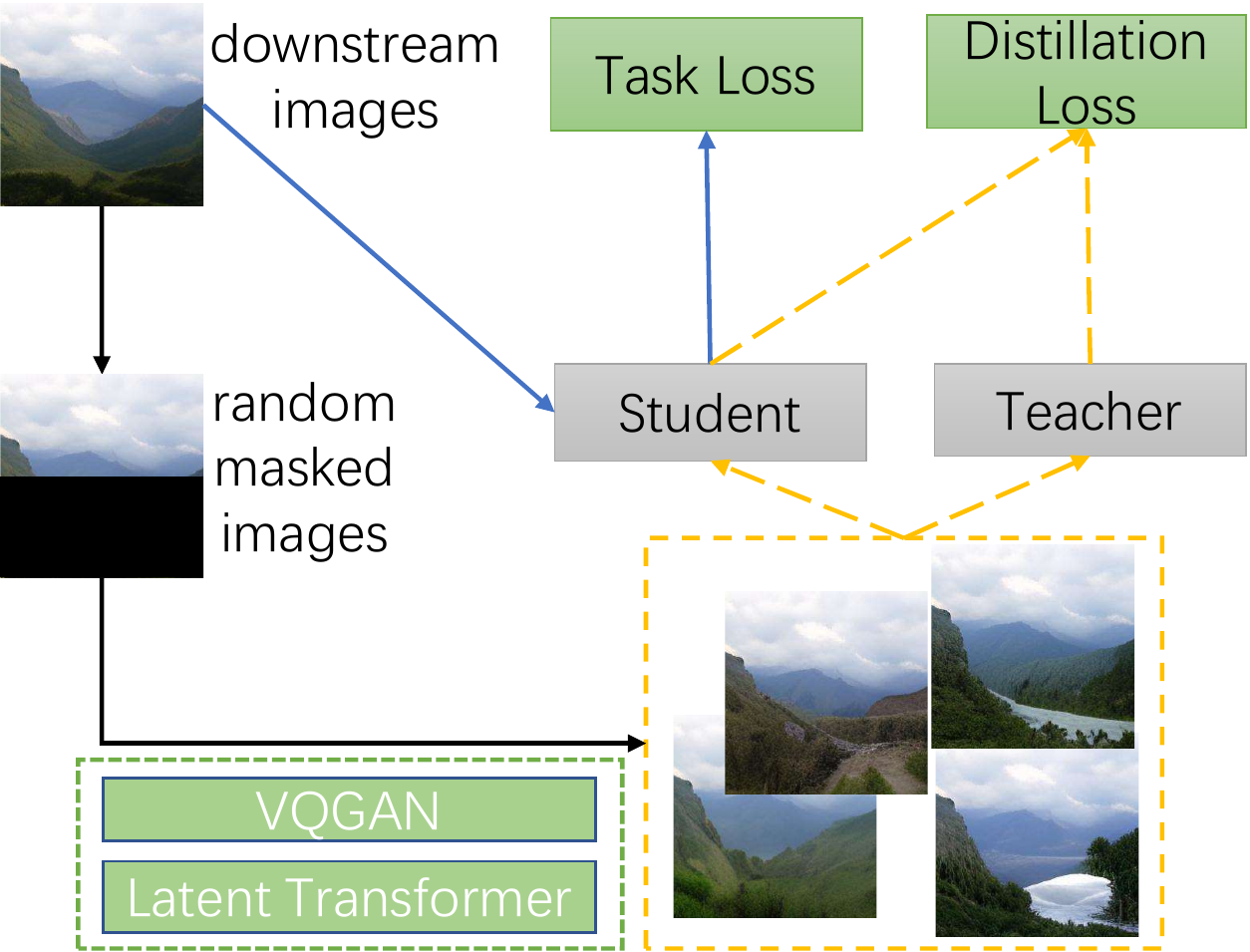}
    \end{center}
    \caption{Overview of DIGG and distillation. Downstream images are randomly masked and then the masked images are completed by the trained VQGAN model and latent transformer with the image prior. The completed images are used for knowledge distillation. Best viewed in color.}
\label{fig:digg.pdf}
\end{figure}

With the VQGAN model and latent transformer, we can generate massive images. In order to generate codebook indices that conform to the downstream data distribution, we can fine-tune the trained latent transformer with downstream data. However, due to the amount of downstream data, it is not easy for the model to be adapted.
We thus first randomly mask some parts of an image (e.g. the bottom half of an image), then encode it by $\mathbf{E}$ and represent it with $\mathbf{Z}$.
The indices of the remaining parts are used as the initial input of the latent transformer, then the indices of masked parts are predicted by the latent transformer in an auto-regressive manner.
This process will perform many times for each downstream image, and the generated image will only be used for knowledge distillation, as illustrated in Figure~\ref{fig:digg.pdf}.
We simply use L2 loss for our experiments.


\subsection{One to Transfer All}
In this section, we will introduce our proposed universal transfer framework: One to Transfer All (OTA), which connects the two methods above. 

We firstly transfer a VFM to a task-specific model by IRF, then perform knowledge distillation to transfer information from the large task-specific model to a small deployed model with images generated by DIGG. Finally, the distilled model will be fine-tuned on the downstream dataset.
The last fine-tuning can be merged to the distillation process, but it will introduce another hyper-parameter to adjust. So we bring it to an independent phase to avoid annoying hyper-parameter searching, which nearly has no influence on final performance.
Note that foundation models are getting bigger than bigger, most end users may not have sufficient resource to fine-tune them.
So for users who own sufficient resources, it's better to perform IRF first then distill it to a small model with data produced by DIGG.
Otherwise, performing distillation at first then using IRF to transfer can also reach fairly well performance. It will be analyzed in Section~\ref{sec:irf-digg-digg-irf}.




\section{Experiments}

In this part, we exhibit the experimental results of our methods. We conduct our methods on different downstream tasks, e.g. classification, detection, segmentation, and compare it with many commonly used transfer methods.
All results in this section are conducted with an officially released VFM: BiT-M-R50x1~\cite{kolesnikov2020big}, if not explicitly noted. 
To construct few data setting, for each downstream task, we randomly sample 10\% data from the full dataset to show the strength of our approach on few data transfer scenario.

\subsection{IRF on Downstream Tasks}

\subsubsection{Implementation Details}
\label{sec:irf_impl}

As shown in Figure~\ref{fig:transfer-vp-pipeline}, our transfer pipeline has 4 stages. For Stage 1 and Stage 2, we train the VQGAN to encode upstream information following the settings in~\cite{esser2021taming}. For Stage 3, which uses images re-represented by VQGAN, we freeze the model's backbone and only update task-specific parameters. We train the model for 5000 steps with a batch size of 64. The optimizer is SGD with Nesterov momentum and a momentum parameter of 0.9 is used in this stage. The initial learning rate is grid searched log-uniformly from the range [1, 1e-3] and follows a multi-step decay schedule with weight decay 1e-5. Any input image is resized to 224, and a random square crop from the resized image is the only data augmentation used during training. For Stage 4, all model parameters are fine-tuned with original downstream task data to calibrate the model precision. The initial learning rate is chosen from a grid of 4 logarithmically spaced values in the range [1e-2, 1e-5], and the weight decay is similarly chosen from a grid of 3 logarithmically spaced values between 1e-3 and 1e-5. The other hyper-parameters and training steps are the same as those in Stage 3. For linear probe and fine-tuning, we use the same settings as Stage 4 except for using 10000 training steps to make a fair comparison. 

Specially, for Swin-Transformer, we use AdamW~\cite{loshchilov2017decoupled} as our optimizer, with $\beta1 = 0.9$, $\beta2 = 0.999$ and weight decay 1e-8. Other hyper-parameters are the same with CNN-based model's mentioned above.

In Section~\ref{sec:sota-comparison}, we conduct experiments on the 100\% dataset to compare our results with state-of-the-art methods. For 100\% data setting, we follow the same hyper-parameter strategy as described above except for more training steps: 10000 iterations in Stage 3 and Stage 4 each. 

\subsubsection{Classification}
For the classification task, we conduct IRF on 17 classification tasks and compare the results with commonly used fine-tuning and linear-probe methods in Table~\ref{tab:cls-downstream-transfer} with a 10\% data setting. The task specific parameters updated for IRF Stage 3 is the last fully connected layer.

As the results exhibited in Table~\ref{tab:cls-downstream-transfer}, our transfer method shows an evident performance improvement compared with linear probe and fine-tuning. For an average accuracy on all tasks, our approach improves 1.9\% by fine-tuning on top-1 accuracy.
In the few data setting of 10\% classification downstream dataset, our IRF method is more effectively utilizing the pre-trained VFM compared with the linear probe or fine-tuning method.
In nearly all downstream dataset, our method outperforms linear probe and fine-tuning.

\begin{table}
\centering
\footnotesize
\begin{tabular}{lccc}
\toprule  
Dataset    &             Linear Probe &                       Fine-tuning &    Ours           \\
\midrule   
CIFAR10 \cite{cifar}&      91.4 &          92.6	       &              \textbf{94.2}	     	 \\
CIFAR100 \cite{cifar}           &       70.6 &     71.3        &              \textbf{73.9}      \\
Flowers \cite{flowers}           &    83.4     &     78.6        &              \textbf{84.6}      \\
Food101 \cite{food}           &        70.8 &     71.2        &              \textbf{74.4}      \\
Pets \cite{pets}           &           74.6 &     80.2        &              \textbf{80.4}      \\
SUN397 \cite{sun}           &         42.7 &     42.6        &              \textbf{44.8}      \\
Stanford Cars \cite{stanfordcar}           &  19.9 &     \textbf{20.2}        &              19.7      \\
DTD \cite{DTD}           &        52.3 &         54.7        &              \textbf{57.1}      \\
Caltech101 \cite{Caltech101}           &    72.6 &      75.0        &              \textbf{75.8}      \\
AirCraft \cite{aircraft}           &      19.4 &      22.7        &              \textbf{25.9}      \\
svhn \cite{svhn}           &         57.5 &       89.8        &              \textbf{92.8}      \\
eurosat \cite{eurosat}           &       93.2 &      95.2        &              \textbf{95.5}      \\
resisc45 \cite{resisc45}           &      81.3 &      85.2        &              \textbf{87.0}      \\
retinopathy \cite{retinopathy}           &    66.5 &     73.8        &              \textbf{75.2}      \\
fer2013 \cite{fer2013}           &      48.1 &       54.4        &              \textbf{54.7}      \\
ucf101 \cite{ucf101}           &       56.7 &       \textbf{59.8}        &              59.6      \\
gtsrb \cite{GTSRB}           &        73.3 &        89.4       &              \textbf{93.5}      \\
\midrule
Average     &     63.2 &     68.1  &   \textbf{70.0}    \\
\bottomrule
\end{tabular}
\caption{\label{tab:cls-downstream-transfer}  Performance of our approach on classification downstream tasks with Bit-M-R50x1. }
\end{table}

\subsubsection{Object Detection and Semantic Segmentation}
To further prove the generality of our approach, we evaluate our IRF on the object detection task and semantic segmentation task. For object detection downstream tasks, we use Mask-RCNN~\cite{he2017mask} as our detection method following its original settings.
As for segmentation, DeepLabv3\cite{chen2017rethinkingdeeplabv3} is used. 
We use Bit-M-R50x1 as the backbone model. Task head that updated in our IRF Stage 3 is the network after backbone part, i.e. FPN, RPN, cls/det head in detection and ASPP, cls head in segmentation. 

Table~\ref{tab:det-downstream-transfer} exhibits the result. Our method achieves 0.8\% improvement on COCO detection, and rather significant improvements on Pascal VOC, i.e. 1.6\% for detection and 6.6\% for segmentation. These results show the generality of our method on different tasks.

\begin{table}
\centering
\footnotesize
\begin{tabular}{ccccc}
\toprule  
 Downstream Task & Dataset    &             LP &                       FT &    Ours           \\
\midrule   
Detection & COCO \cite{COCO}&     15.8  &       23.5        &       24.3 \\
Detection & Pascal VOC \cite{pascal}           &   68.4  &  68.3  &  70.0                  \\
\midrule
Segmentation & Pascal VOC \cite{pascal}           &        53.4 &     58.8        &             65.4      \\
\bottomrule
\end{tabular}
\caption{\label{tab:det-downstream-transfer}  Performance of our approach on object detection and semantic segmentation. We show mAP results on object detection and mIOU results on semantic segmentation.}
\end{table}

\subsubsection{Comparison with State-of-the-arts}
\label{sec:sota-comparison}
We further compare results with advanced transfer methods in recent studies of transfer learning, including BSS\cite{chen2019catastrophic}, DELTA\cite{li2019delta}, SpotTune\cite{guo2019spottune}, PtR\cite{zhong2020regularizing}, Adafilter\cite{guo2020adafilter} and StochNorm\cite{kou2020stochastic}.
We follow their data settings and network structure. We apply IRF on ImageNet pre-trained ResNet-50 network for a fair comparison.
Due to different optimization settings among these methods, we report relative improvements between their method and  baseline in their paper.
We do observe that our transfer method outperforms almost all of the advanced transfer methods, indicating the strength of our proposed methods.

\begin{table}[t]
    \centering
    \resizebox{1.0\linewidth}{!}{
    \begin{tabular}{l|cccc}
    \toprule
    Transfer Method & Flowers($\varDelta$) & CUBS($\varDelta$) & AirCraft($\varDelta$) & Stanford Cars($\varDelta$) \\
    \midrule
    BSS \cite{chen2019catastrophic}   &   -  & $+$0.84 & $+$0.35 & $+$0.43 \\
    DEALTA \cite{li2019delta}         &   -  & $+$0.62 & $-$0.69 & $-$0.88 \\
    SpotTune \cite{guo2019spottune}   & $+$2.93 & $+$2.17        & -           & \textbf{$+$2.66}\\
    PtR \cite{zhong2020regularizing}  & $+$0.80 & $+$1.60        & -           & -    \\
    Adafilter \cite{guo2020adafilter} & -    &   -          &$+$2.82        & -    \\
    StochNorm \cite{kou2020stochastic}& -    & $+$1.57 & $+$0.53 & $+$1.57 \\
    \midrule
    ours                              &  \textbf{$+$3.33} &     \textbf{$+$3.36}         &  \textbf{$+$2.87}       & $+$1.26 \\
    \bottomrule
    \end{tabular}
    }
    \caption{Comparison results with state-of-the-art transferring methods with ImageNet pre-trained ResNet-50. We show the relative improvement between methods and baselines.
    }
    \label{tab:advanced_finetune}
\end{table}

\subsubsection{Ablation on IRF}

\begin{table*}[h]
    \centering
    \resizebox{0.9\linewidth}{!}{%
    \begin{tabular}{lll|cccccc|c}
    \toprule
    Setting & Delivering Data & Calibration Data & CIFAR10 & CIFAR100 & Flowers & Food101 & Pets & SUN397 & AVG\\
    \midrule
    Finetune  & - & Original & 92.6& 71.3& 78.6 & 71.2& 80.2& 42.6 & 72.8\\
    Linear Probe  & Original &  - & 91.4& 70.6& 83.4& 70.8& 74.6& 42.7 & 72.3\\
    \midrule
    (a)  & Original &  Original & 91.2& 71.5& 83.4& 71.5& 79.2& 42.9 & 73.3\\
    (b)  & Official Rec &  Original & 92.4& 71.9& 83.5& 72.6& 79.8&43.1 &73.9\\
    (c) & Rec & Rec & 91.6& 70.3& 80.7& 69.5& 76.4& 42.9 & 71.9\\
    (d) Test on Rec Val& Rec & Rec & 94.0& 72.9& 85.1& 73.1& 80.2& 44.8 & 75.0\\
    (e)  & - &  Rec + Original & 92.7& 71.6& 79.3& 71.5& 80.1& 42.4 & 73.0\\
    (f)  & Rec + Original     &  Rec + Original & 93.3& 72.2& 83.8& 72.0& 79.9& 43.1 & 74.1\\
    \midrule
    Ours  & Rec &  Original & 94.2 & 73.9& 84.6 & 74.4& 80.4& 44.8 & \textbf{75.4}\\
    \bottomrule
    \end{tabular}

    }
    \caption{\label{tab:codebook_ablation}Performance of different designs in stage 3 and stage 4 of IRF.}
\end{table*}

In this section, we show the effectiveness of our Image Re-representation Fine-tuning (IRF) method and provide detailed experiments on each stage of the IRF pipeline.

\textbf{Effectiveness of Stage 3 Delivering.} We compare our method with not using re-represented images, which means we use the original downstream task images in this stage, and use the same optimization strategy in stage 4.
We show results with classification downstream tasks in setting (a) of Table~\ref{tab:codebook_ablation}. Results show that delivering stage plays an important role in the IRF pipeline. Using re-represented images will bring an average improvement of 2.1\%.

We also compare results with using an officially released VQGAN model which is trained with ImageNet1k~\cite{esser2021taming} in Table~\ref{tab:codebook_ablation} setting (b).
Due to the domain gap introduced by different training datasets of VQGAN and VFM, the performance improvement of officially released VQGAN is limited and the average performance is 1.5\% lower than ours.

\begin{figure}
  \begin{center}
   \includegraphics[width=0.9\linewidth]{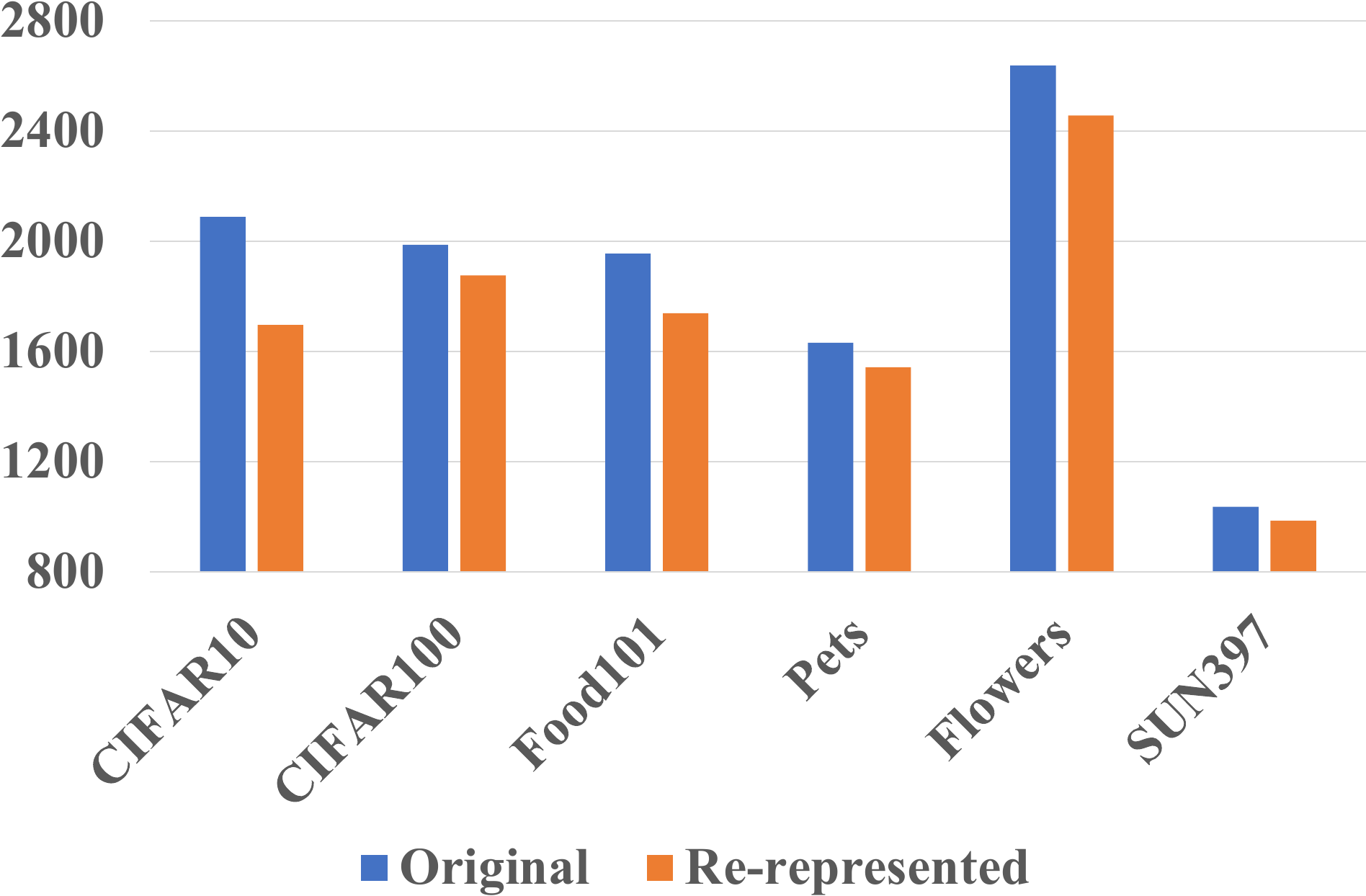}
    \end{center}
    \caption{FD scores comparison of upstream dataset with downstream original dataset and re-represented dataset. FD score lower means two datasets are more similar.}
\label{fig:fd_score}
\end{figure}

\textbf{Re-represented Image VS Original Image.} 
From our intuition in IRF method, the re-represented downstream images are more similar to upstream data and can extract more reliable features with the upstream data pre-trained model. To further demonstrate this, we provide quantitative analysis on the similarity between upstream dataset and reconstructed downstream dataset as well as the similarity between upstream dataset and original downstream dataset. We use FD-score to measure the dataset similarity following the method described in~\cite{deng2021labels}. Figure~\ref{fig:fd_score} exhibits that the FD score between upstream dataset and re-represented dataset is consistently lower, which means that our VQGAN model successfully encodes upstream information and re-represents the original downstream data into a domain that is more similar to upstream data. With the re-represented images, a VFM can extract more reliable features.

\textbf{Effectiveness of Stage 4 Calibration.} In Stage 4 of the IRF pipeline, we calibrate the model with original images to refine the model in the downstream domain. In this part, we show the results of using re-represented images in the calibration stage in setting (c) of Table~\ref{tab:codebook_ablation}. Compared with ours, the performance decreased 3.5\%, which shows the importance of calibration with original downstream data. Furthermore, in setting (d) of Table~\ref{tab:codebook_ablation}, we test the model on the re-represented downstream validation set, which is trained in setting (c).
Results show that the model trained by re-represented images can reach good performance with the same domain but still lower than ours.
We think this is because re-represented images may lose detailed information to some extent.
Besides, using re-represented images for testing may hinder its scalability. So we use original images for optimization in stage 4.

\textbf{Re-represented Image VS Data augmentation.} It is easy to consider a problem that whether using VQGAN to re-represent the image is just a kind of data augmentation. To check this, we mix up the re-represented and original downstream images in Stage 3 and Stage 4 of IRF. Setting (e) and (f) in Table~\ref{tab:codebook_ablation} corresponds to using mixed data with the fine-tuning method and using mixed data with a two-stage fine-tuning process. Results show that the mixture of data indeed brings some improvement compared with simply using original data, but the gain is limited and still 1.3\% lower than our final results.
So it does not only plays a role in data augmentation but also makes the extracted features more reliable to optimize task-specific parameters better.

\subsection{Distillation with DIGG}
\label{sec:exp_digg}

In this section, we conduct experiments of distilling knowledge from large VFMs into a small model with our DIGG approach. We exhibit results on different downstream tasks to validate the effectiveness and generality of our approach. The teacher and student are BiT-M-R50x1 and ResNet-18 by default.

\subsubsection{Implementation Details}
\label{sec:digg_impl}
We use the transferred BiT-M-R50x1 as the teacher model and ResNet-18 as the student model.
We first use the DIGG method to generate 100k images that conform to the downstream distribution if not explicitly noted.
Then the output of the backbone's last layer of the teacher model and the student model are used for distillation, with $L2$ loss function.
Every dataset is distilled for 70 epochs. For the optimization hyper-parameters, we use the SGD optimizer, with a learning rate of 1.0, weight decay 1e-5, momentum 0.9. When finally fine-tuning, we use the same setting mentioned in Section \ref{sec:irf_impl} stage 4.

\subsubsection{DIGG on Multiple Downstream Tasks}

\textbf{Classification.} Table~\ref{tab:diff_vfm_for_framework} shows the effectiveness of our proposed DIGG method on six classification tasks. 
With our DIGG, a ResNet-18 model distilled by BiT-M-R50x1 can achieve 1.3\% higher average accuracy than fine-tuned from ImageNet pre-train. And it is also higher than the normal distillation method by a large margin.

\noindent\textbf{Detection and Segmentation.} Table~\ref{tab:det-seg-digg} shows the performance of applying DIGG on object detection and semantic segmentation. With 400k generated images, performance increases 6.0 and 3.4 in mAP for COCO and VOC detection task compared with using ImageNet pre-trained backbone. Besides, it also shows an improvement of 5.5 in mIOU on VOC for the segmentation task, which further validates the generality of the DIGG method on different downstream tasks. What's more, as shown in the fourth column, distilling with 10\% downstream data results in poor performance, even worse than fine-tuning from the ImageNet pre-trained model, which demonstrates the difficulty of distillation with few data and shows the power of our DIGG approach.

\subsubsection{Influence of Image Distribution on Distillation}
\label{sec:ID_DIGG}

\begin{table*}[ht]
    \centering
    \footnotesize
    
    \begin{tabular}{cc|cccccc|c}
    \toprule
    Setting     &   Distillation data  & CIFAR10 & CIFAR100 & Flowers & Food101 & Pets & SUN397 & Average\\
    \midrule
    (a) &         -            &  89.9   &     67.1    &   67.3     &   66.1    &   83.9   &    33.9 & 68.0 \\
    (b) &   Downstream raw(10\%)   & 90.5      &  61.4       &   39.4   &  68.4   &  24.3   & 24.3  & 51.4 \\
    (c) &   Downstream raw(100\%)  & 93.9   &    74.3    &  87.1  &  77.4   &  77.2   & 45.4  & 75.9\\
    \midrule
    (d) &   Upstream(100k)      &  87.8   &  59.2  &  55.8 &  46.6 &  43.9 & 30.3  & 53.9 \\
    (e) &   Upstream(400k)      &  89.4   &  65.0  & 56.9  & 59.6 & 50.2 & 32.2 & 58.9 \\
    (f) &   DIGG(100k)       & 90.8  &  65.4  &  75.1  & 70.1 & 71.8 & 36.4 & 68.3 \\
    (g) &   DIGG(400k)       & 92.2  &  67.7  &  75.0  & 71.9 & 71.4 & 37.5 & 69.3 \\
    \bottomrule
    \end{tabular}
    
    \caption{Comparison among distillations by different distribution data. The teacher model is BiT-M-R50x1. The student model is ResNet-18 and it is pre-trained by ImageNet only in setting (a). After distillation, the model is further fine-tuned on 10\% downstream data.}
    
    \label{tab:digg_ablation}
\end{table*}

\begin{table}
\footnotesize
\centering
\begin{tabular}{cccccc}
\toprule  
 Downstream Task & Dataset    &             IN-PT &  KD &  Ours           \\
\midrule   
Detection & COCO \cite{COCO}               &       14.0& 9.4& 20.0      \\
Detection & Pascal VOC \cite{pascal}       &      55.1&   28.2&  58.5   \\
\midrule
Segmentation & Pascal VOC \cite{pascal}    &      46.8&   26.8&  52.3   \\
\bottomrule
\end{tabular}
\caption{\label{tab:det-seg-digg} Performance of DIGG on object detection and semantic segmentation. BiT-M-R50x1 and ResNet-18 is used as teacher and student. IN-PT refers to ImageNet pre-trained ResNet-18 fine-tuned on 10\% downstream dataset. KD refers to a ImageNet pre-trained ResNet-18 distilling with 10\% downstream data. We show mAP result for detection and mIOU result for segmentation. 
}
\end{table}

DIGG can be seen as generating images that follow a certain distribution. In this part, we will analyze the influence of image distribution on distillation.

When performing KD, the distillation data with different distributions will make the student model focus on different parts of a complex function modeled by a neural network.
It is worthy to explore what kind of distribution of distillation data will make the student model benefit more when transferring to downstream tasks. 


The results are shown in Table \ref{tab:digg_ablation}. 
All experiments are first distilled by the listed dataset then fine-tuned on the 10\% labeled downstream data.
Setting (a), (b), (c) are our baselines. 
In setting (a), the student is pre-trained on ImageNet dataset and directly fine-tuned on 10\% downstream data without distillation.
In setting (b), the student is distilled from scratch by 10\% original downstream data and then fine-tuned.
In setting (c), we use 100\% downstream data to distill and it can be seen as a strong baseline since it leaks 90\% data that will not be used in other settings.
From these three baselines, ImageNet pre-train can largely boost performance compared with setting (b), but still inferior to 100\% downstream data distillation, which indicates distilled knowledge from VFM is more effective than ImageNet pre-train.
From setting (d) to (g), we compare the influence of distribution and amount of distillation data.
Upstream distribution data is directly sampled, and downstream distribution data is generated by our DIGG approach.
We can see that distillation data from downstream distribution achieves higher performance than it from upstream data distribution, even with less data.
With 400k distillation data generated by DIGG, the performance can exceed ImageNet Pre-train by 1.3\%.
We think the performance will still boost when increasing the generated distillation data by DIGG.

\subsubsection{DIGG compared with Deep Inversion}
In order to compare the pros and cons of our DIGG with other image synthesis methods, we follow the Deep Inversion method mentioned in \cite{Yin_2020_CVPR} to generate images conforming to downstream distribution. For deep inversion, we fine-tune a BiT-M-50x1 on 10\% data  for generation.
We generate 100k images for both methods to compare their results on distillation.
For distillation, we both use BiT-M-R50x1 as the teacher to transfer knowledge to ResNet-18 with different pseudo data, following the same settings mentioned in Section \ref{sec:ID_DIGG} and reporting the top-1 accuracy of the student model.
As shown in Table~\ref{tab:digg_cmp_di}, using pseudo data generated by our DIGG method can achieve an average improvement of 2.7\% than Deep Inversion.

\begin{table}[h]
\centering
\resizebox{0.95\linewidth}{!}{
    \begin{tabular}{l|cccccc|c}
    \toprule  
    Dataset  & CIFAR10 & CIFAR100 &  Flowers & Food101 & Pets & SUN397 & AVG \\
    \midrule   
    DI\cite{Yin_2020_CVPR}  & 90.0 & 61.1 & 73.1 & 66.0 & 70.5 & 33.1 & 65.6 \\
    DIGG & 90.8 & 65.4& 75.1 & 70.1 & 71.8 & 36.4 & 68.3 \\
    \bottomrule
    \end{tabular}
}
\caption{\label{tab:digg_cmp_di} Knowledge transfer from BiT-M-R50x1 to ResNet-18 with data generated from different methods on few data setting.}
\end{table}

\subsection{Image Visualization by DIGG}
\begin{figure*}
  \begin{center}
   \includegraphics[width=1.0\linewidth]{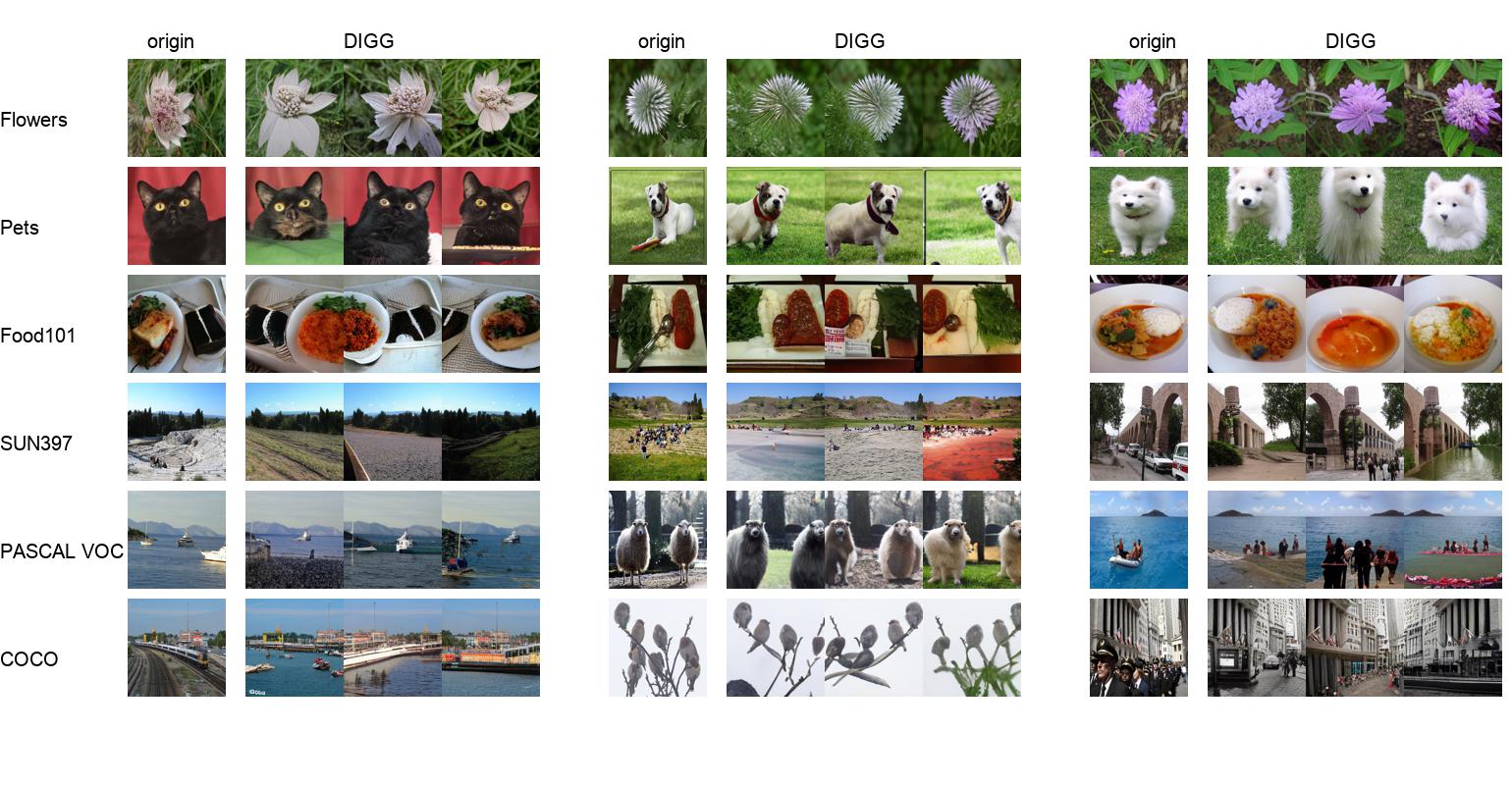}
    \end{center}
    \vspace{-2em}
    \caption{Pseudo Images generated by our DIGG method on different datasets. There are three groups of images in each data set. For each group, the leftmost column is the original image, and the rest are different images generated by the our DIGG method from the same original image.}
\label{fig:digg_img}
\end{figure*}

In Figure \ref{fig:digg_img}, we show generated images by our DIGG method on different downstream datasets. It can be seen that pseudo images generated by our DIGG method can introduce more diversity but still follow the main content and style of the downstream.

\subsection{Overall Results of OTA}

In this section, we show the experimental results of our overall framework OTA.
Methods are evaluated on the same six classification datasets if not explicitly noted, i.e. CIFAR10, CIFAR100, Flowers, Food101, Pets, and SUN397, and the average accuracy of these datasets are reported.
We only use 100k images for distillation by DIGG in this part.

\subsubsection{Transfer from Different VFMs}
In this part, we use different VFMs to validate our framework, including Swin-Transformer~\cite{liu2021swin}, SWAV~\cite{caron2021unsupervised}, CLIP R50~\cite{radford2021learning}, BiT-M-R50x1~\cite{kolesnikov2020big} and BiT-M-R101x1~\cite{kolesnikov2020big} in Table~\ref{tab:diff_vfm_for_framework}. 
The first two row blocks compare the results with fine-tuning and our IRF approach. For each VFMs, our IRF approach shows its superiority over fine-tuning. 
The last two row blocks compare the results of using vanilla KD and using our IRF-DIGG approach in the KD process. Due to the 10\% few data setting, vanilla KD can only achieve a bad performance, indicating the difficulty of transferring a large VFM model into a small model under few data scenarios. For IRF-DIGG, it shows prominent improvements with all VFM models which further prove the effectiveness of our approach.
We also provide ImageNet pre-trained models for comparison, with our OTA, the average accuracy of downstream tasks can boost up to 2.9\% than ImageNet pre-trained ResNet-18.
It demonstrates that even no strong VFM with a small backbone exists, like ResNet-18, our OTA can still provide a better choice than the ImageNet pre-trained model.
Note that the Swin-Transformer-L reach highest fine-tuning and IRF results, but the performance of KD and IRF-DIGG is not satisfactory. We think this problem is due to the KD method itself between the Swin-Transformer and ResNet-18 and it will be left as our future work.

\begin{table}[]
\centering
\resizebox{1.0\linewidth}{!}{%
\begin{tabular}{ll|l|lllll}
\toprule
                             &          & \rotatebox{90}{ResNet-18} & 
                             \rotatebox{90}{Swin-L} &
                             \rotatebox{90}{SWAV-R50} & \rotatebox{90}{CLIP-R50} & \rotatebox{90}{BiT-M-R50x1} & \rotatebox{90}{BiT-M-R101x1} \\
\midrule
\multirow{7}{*}{\rotatebox{90}{Fine-tuning}} & CIFAR10  &   89.9  & 98.2 &  92.3   &      82.0       &      92.6    &      94.5    \\
                             & CIFAR100 & 67.1 & 89.7 & 69.0 &     52.8        &       71.3   &      77.5    \\
                             & Flowers  & 67.4 & 96.9 & 66.3 &     47.3        &       78.6   &      85.2    \\
                             & Food101  & 66.0 & 89.0 & 73.2 &     71.2        &       71.2   &      74.5    \\
                             & Pets     & 83.8 & 90.4 & 71.6 &     56.0        &       80.2   &      83.2    \\
                             & SUN397   & 33.9 & 64.4 & 38.2 &     51.6        &       42.6   &      50.1    \\
                             & AVG      & 68.0 & 88.1 & 68.4 &    60.1         &        72.8  &      77.5    \\
\midrule
\multirow{7}{*}{\rotatebox{90}{IRF}}         & CIFAR10  &     -     & 98.8 & 93.2 & 86.0 &      94.2    &      95.9    \\
                             & CIFAR100 &    -       &  90.9 &  74.1 &  59.2&        73.9  &      78.9    \\
                             & Flowers  &    -     &  98.9 &  69.9 &  59.1&        84.6  &      87.6    \\
                             & Food101  &      -    &  89.9 &  74.8 &  71.2&        74.4  &     76.5     \\
                             & Pets     &    -     &  92.4 &  72.5 &  66.6&        80.4  &      84.4    \\
                             & SUN397   &      -      &  69.4 &  45.1 &  52.8&         44.8 &      51.2    \\
                             & AVG      &     -      &  90.1 &  71.7 &  65.8&      75.4   &     79.1 \\ 
\midrule
\multirow{7}{*}{\rotatebox{90}{KD}} & CIFAR10          &     -   & 85.6 & 71.4 & 85.7 & 90.5 & 88.0       \\
                             & CIFAR100  &   -  & 52.3 & 28.3 & 53.1 & 61.4 & 56.3   \\
                             & Flowers  &   - & 39.5 & 18.4 & 37.1 & 39.4 & 38.4    \\
                             & Food101  &    - & 56.1 & 38.3 & 64.7 & 68.4 & 67.9   \\
                             & Pets     &    -  & 20.7 & 26.0 & 41.7 & 24.3 & 52.3   \\
                             & SUN397   &    -  & 15.3 & 8.4  & 13.1 & 24.3 & 19.1   \\
                             & AVG      &    -  & 44.9 & 31.8 & 49.4 & 51.4 & 53.7   \\
\midrule
\multirow{7}{*}{\rotatebox{90}{IRF-DIGG}}         & CIFAR10  &-& 88.1 & 90.1 & 87.9 & 92.7 & 91.5\\
                             & CIFAR100 & -  & 65.5 & 68.7 & 61.6 & 66.4 & 66.8 \\
                             & Flowers  & -& 70.5 & 62.2 & 57.8 & 81.2 & 76.9 \\
                             & Food101  & -  & 69.5 & 68.2 & 63.6 & 70.5 & 72.5 \\
                             & Pets     & -  & 73.4 & 72.1 & 62.4 & 74.6 & 79.5 \\
                             & SUN397   & -  & 29.3 & 33.2 & 31.3 & 38.2 & 38.7 \\
                             & AVG      & -& 66.1 & 65.7 & 60.8 & 70.5 & 70.9 \\        
\bottomrule
\end{tabular}
}
\caption{Detailed results of fine-tuning, IRF, KD and DIGG. The student for all VFMs is ResNet-18.}
\label{tab:diff_vfm_for_framework}
\end{table}

\subsubsection{IRF-DIGG VS DIGG-IRF}
\label{sec:irf-digg-digg-irf}
As discussed in Section~\ref{sec:digg}, it is better to perform IRF and then apply DIGG. But users may not have sufficient computing resources to directly apply IRF on a large VFM. As a result, they can perform distillation at first and then use IRF to transfer. In this part, by using VFM BiT-M-R50x1, we compare the performance of IRF-DIGG and DIGG-IRF in Table \ref{tab:irf-digg-digg-irf-ablation}.
From the table we can see that using DIGG-generated images for KD first and then performing IRF can also achieve comparable results, and can reach higher performance than ImageNet pre-train in some datasets. Besides, applying IRF first and then applying DIGG consistently achieves higher results because the IRF model is adapted to downstream domain and it is a better teacher for student model to learn from according to downstream data.

\begin{table}
\centering
\footnotesize
\begin{tabular}{l|c|cc}
\toprule  
ResNet-18  & IN-PT &  IRF-DIGG  &     DIGG-IRF   \\
\midrule   
CIFAR10 \cite{cifar}       &  89.9  &  92.7     &  91.1    \\
CIFAR100 \cite{cifar}      &  67.1  &  66.4     &  66.4   \\
Flowers \cite{flowers}       &  67.3 &  81.2     &  77.7   \\
Food101 \cite{food}       &  66.1  &  70.5     &  70.4   \\
Pets \cite{pets}          &  83.9  &  74.6     &  71.3   \\
SUN397 \cite{sun}        &  33.9  &  38.2     &  40.0   \\
\midrule   
Average               &  68.0  &  70.5     &  69.5    \\
\bottomrule
\end{tabular}
\caption{\label{tab:irf-digg-digg-irf-ablation} Comparison results between IRF-DIGG and DIGG-IRF. IN-PT means ResNet-18 ImageNet pre-train results.}
\end{table}

\subsection{Optimization after IRF-DIGG}
In our OTA pipeline, we simply use fine-tuning after the model is firstly optimized by IRF then distilled by DIGG by default.
It is possible that even the model is small, it may be still hard to jointly optimize the small backbone and task-specific parameters in few data setting.
So we make an additional experiments to compare the two choices. 
In Table \ref{tab:opt-after-digg}, we follow up with firstly linear probe then fine-tuning (LP+FT) strategy and find the LP+FT strategy only has marginally gain in performance. So we choose fine-tuning(FT) after IRF-DIGG.

\subsubsection{Transfer on VTAB Benchmark}
\label{sec:vtab-benchmark}

\begin{table*}[h]
    \centering
    \resizebox{1.0\linewidth}{!}{%
    \begin{tabular}{l|lllllll|llll|llllllll|l}
    \toprule
     & \rotatebox{90}{Caltech101} & \rotatebox{90}{CIFAR100} & \rotatebox{90}{DTD} & \rotatebox{90}{Flowers102} & \rotatebox{90}{Pets} & \rotatebox{90}{SUN397} & \rotatebox{90}{SVHN} & \rotatebox{90}{Camelyon} & \rotatebox{90}{Eurosat} & \rotatebox{90}{Resisc45} & \rotatebox{90}{Retinopathy} & \rotatebox{90}{Clever-Count} & \rotatebox{90}{Clever-Dist} & \rotatebox{90}{DMLab} & \rotatebox{90}{dSpr-Loc} & \rotatebox{90}{dSpr-Ori} & \rotatebox{90}{KITTI-Dist} & \rotatebox{90}{sNORB-Azim} & \rotatebox{90}{sNORB-Elev} & \rotatebox{90}{AVG}\\
    \midrule
    ResNet50-Finetune\dag & 89.8 & 54.6 & 65.6 & 88.4 & 89.1 & 34.5 & 86.3 & 79.7 & 95.3 & 81.0 & 72.6 & 41.8 & 52.5 & 42.7 & 81.0 & 47.3 & 75.3 & 32.6 & 35.8 & 65.6 \\
    ResNet50-IRF & 90.5 & 55.3 & 62.7 & 92.0 & 91.3 & 37.4 & 85.3 & 84.4 & 95.8 & 82.9 & 73.0 & 52.6 & 56.6 & 44.2& 86.9& 50.2& 76.6& 20.9& 37.5 & \textbf{67.1}\\
    \midrule
    ResNet18-Finetune & 85.9 & 45.1 & 55.8 & 90.4 & 88.0 & 27.0 & 81.2 & 74.7 & 80.1 & 80.9 & 72.3 & 39.5 & 50.2 & 42.2 & 77.2& 51.3& 72.2& 19.9& 35.1 & 61.5\\
    ResNet18-OTA & 83.2 & 49.7 & 56.1 & 90.5 & 86.6 & 24.7 & 86.8 & 83.9 & 95.7 & 83.6 & 73.7 & 65.4 & 57.9 & 41.9 & 49.6 & 60.4 & 71.8 & 22.2 & 40.9 & \textbf{64.5} \\
    \bottomrule
    \end{tabular}

    }
    \caption{Our transfer method on VTAB-1K benchmark. \dag refers to results directly copied from VTAB paper~\cite{zhai2019large}. ResNet18-Finetune refers to fine-tuning results on VTAB-1K with ImageNet pre-trained ResNet18 model follow lightweight hyper-parameter modes. ResNet18-OTA uses ResNet50 as teacher model, and 100k images for distillation by DIGG.}
    \label{tab:vtab_benckmark}
\end{table*}

To further assess the generality of our transfer approach, we evaluate our approach on the Visual Task Adaptation Benchmark(VTAB)~\cite{zhai2019large}, which consists of 19 different visual tasks including three groups: \textit{natural}, \textit{specialized} and \textit{structured}. We follow the lightweight hyper-parameter modes as described in VTAB and show the results in Table~\ref{tab:vtab_benckmark}. For an ImageNet pre-trained ResNet-50, our IRF approach achieves an additional 1.5\% in average accuracy over 19 datasets compared with fine-tuning. We also evaluate OTA pipeline on VTAB-1K benchmark by transferring ResNet-50 into a small model, ResNet-18. Comparison results are shown on the last two lines in Table~\ref{tab:vtab_benckmark}, our OTA approach achieves 3.0\% improvement in average compared with fine-tuning from ImageNet pre-trained model.

\begin{table}
\centering
\footnotesize
\begin{tabular}{l|cc}
\toprule
ResNet-18  & IRF-DIGG-FT  &  IRF-DIGG-LP+FT   \\
\midrule
CIFAR10   &  92.7     &  92.4   \\
CIFAR100  &  66.4     &  66.8   \\
Flowers   &  81.2     &  80.8   \\
Food101   &  70.5     &  70.6   \\
Pets      &  74.6     &  74.2   \\
SUN397    &  38.2     &  39.0   \\
\midrule
Average   &  70.5     &  70.6   \\
\bottomrule
\end{tabular}
\caption{\label{tab:opt-after-digg} Comparison of different optimization methods after IRF-DIGG.}
\end{table}

\section{Conclusion}
We proposed One to Transfer All (OTA), a universal transfer framework to transfer knowledge from any foundation model to any downstream deployed models with few data.
We first proposed Image Re-representation Fine-tuning (IRF) to transfer a vision foundation model to a task-specific model, then distilled knowledge from the task-specific model to a deployed model with data produced by Downstream Image-Guided Generation (DIGG). 
The framework has no dependency on foundation models and downstream tasks.
Exhaustive experiments validate its effectiveness, especially with few data.
Besides, OTA also provides an easy way for vision foundation model researchers to release their upstream information for better downstream transferring and not leaking their data.
We hope OTA will push the application of the vision foundation model a further step.


{\small
\bibliographystyle{ieee_fullname}
\bibliography{egbib}

\begin{thebibliography}{10}\itemsep=-1pt

\bibitem{retinopathy}
Michael~David Abr{\`a}moff, Yiyue Lou, Ali Erginay, Warren Clarida, Ryan
  Amelon, James~C Folk, and Meindert Niemeijer.
\newblock Improved automated detection of diabetic retinopathy on a publicly
  available dataset through integration of deep learning.
\newblock {\em Investigative ophthalmology \& visual science},
  57(13):5200--5206, 2016.

\bibitem{ahmed2021unsupervised}
Sk~Miraj Ahmed, Dripta~S Raychaudhuri, Sujoy Paul, Samet Oymak, and Amit~K
  Roy-Chowdhury.
\newblock Unsupervised multi-source domain adaptation without access to source
  data.
\newblock In {\em Proceedings of the IEEE/CVF Conference on Computer Vision and
  Pattern Recognition}, pages 10103--10112, 2021.

\bibitem{bao2021beit}
Hangbo Bao, Li Dong, and Furu Wei.
\newblock Beit: Bert pre-training of image transformers, 2021.

\bibitem{bommasani2021opportunities}
Rishi Bommasani, Drew~A Hudson, Ehsan Adeli, Russ Altman, Simran Arora, Sydney
  von Arx, Michael~S Bernstein, Jeannette Bohg, Antoine Bosselut, Emma
  Brunskill, et~al.
\newblock On the opportunities and risks of foundation models.
\newblock {\em arXiv preprint arXiv:2108.07258}, 2021.

\bibitem{food}
Lukas Bossard, Matthieu Guillaumin, and Luc Van~Gool.
\newblock Food-101 -- mining discriminative components with random forests.
\newblock In {\em ECCV}, 2014.

\bibitem{brown2020language}
Tom~B Brown, Benjamin Mann, Nick Ryder, Melanie Subbiah, Jared Kaplan, Prafulla
  Dhariwal, Arvind Neelakantan, Pranav Shyam, Girish Sastry, Amanda Askell,
  et~al.
\newblock Language models are few-shot learners.
\newblock {\em arXiv preprint arXiv:2005.14165}, 2020.

\bibitem{cao2018partial}
Zhangjie Cao, Mingsheng Long, Jianmin Wang, and Michael~I Jordan.
\newblock Partial transfer learning with selective adversarial networks.
\newblock In {\em Proceedings of the IEEE conference on computer vision and
  pattern recognition}, pages 2724--2732, 2018.

\bibitem{caron2021unsupervised}
Mathilde Caron, Ishan Misra, Julien Mairal, Priya Goyal, Piotr Bojanowski, and
  Armand Joulin.
\newblock Unsupervised learning of visual features by contrasting cluster
  assignments, 2021.

\bibitem{chen2017rethinkingdeeplabv3}
Liang-Chieh Chen, George Papandreou, Florian Schroff, and Hartwig Adam.
\newblock Rethinking atrous convolution for semantic image segmentation.
\newblock {\em arXiv preprint arXiv:1706.05587}, 2017.

\bibitem{pmlr-v119-chen20s}
Mark Chen, Alec Radford, Rewon Child, Jeffrey Wu, Heewoo Jun, David Luan, and
  Ilya Sutskever.
\newblock Generative pretraining from pixels.
\newblock In Hal~Daumé III and Aarti Singh, editors, {\em Proceedings of the
  37th International Conference on Machine Learning}, volume 119 of {\em
  Proceedings of Machine Learning Research}, pages 1691--1703. PMLR, 13--18 Jul
  2020.

\bibitem{chen2020simple}
Ting Chen, Simon Kornblith, Mohammad Norouzi, and Geoffrey Hinton.
\newblock A simple framework for contrastive learning of visual
  representations.
\newblock In {\em International conference on machine learning}, pages
  1597--1607. PMLR, 2020.

\bibitem{chen2019catastrophic}
Xinyang Chen, Sinan Wang, Bo Fu, Mingsheng Long, and Jianmin Wang.
\newblock Catastrophic forgetting meets negative transfer: Batch spectral
  shrinkage for safe transfer learning.
\newblock 2019.

\bibitem{resisc45}
Gong Cheng, Junwei Han, and Xiaoqiang Lu.
\newblock Remote sensing image scene classification: Benchmark and state of the
  art.
\newblock {\em Proceedings of the IEEE}, 105(10):1865--1883, 2017.

\bibitem{chronopoulou2019embarrassingly}
Alexandra Chronopoulou, Christos Baziotis, and Alexandros Potamianos.
\newblock An embarrassingly simple approach for transfer learning from
  pretrained language models.
\newblock {\em arXiv preprint arXiv:1902.10547}, 2019.

\bibitem{DTD}
M. Cimpoi, S. Maji, I. Kokkinos, S. Mohamed, , and A. Vedaldi.
\newblock Describing textures in the wild.
\newblock In {\em CVPR}, 2014.

\bibitem{courty2016optimal}
Nicolas Courty, R{\'e}mi Flamary, Devis Tuia, and Alain Rakotomamonjy.
\newblock Optimal transport for domain adaptation.
\newblock {\em IEEE transactions on pattern analysis and machine intelligence},
  39(9):1853--1865, 2016.

\bibitem{deng2021labels}
Weijian Deng and Liang Zheng.
\newblock Are labels always necessary for classifier accuracy evaluation?,
  2021.

\bibitem{devlin2018bert}
Jacob Devlin, Ming-Wei Chang, Kenton Lee, and Kristina Toutanova.
\newblock Bert: Pre-training of deep bidirectional transformers for language
  understanding.
\newblock {\em arXiv preprint arXiv:1810.04805}, 2018.

\bibitem{doersch2020crosstransformers}
Carl Doersch, Ankush Gupta, and Andrew Zisserman.
\newblock Crosstransformers: spatially-aware few-shot transfer.
\newblock {\em arXiv preprint arXiv:2007.11498}, 2020.

\bibitem{enderich2021holistic}
Lukas Enderich, Fabian Timm, and Wolfram Burgard.
\newblock Holistic filter pruning for efficient deep neural networks.
\newblock In {\em Proceedings of the IEEE/CVF Winter Conference on Applications
  of Computer Vision}, pages 2596--2605, 2021.

\bibitem{ericsson2021well}
Linus Ericsson, Henry Gouk, and Timothy~M Hospedales.
\newblock How well do self-supervised models transfer?
\newblock In {\em Proceedings of the IEEE/CVF Conference on Computer Vision and
  Pattern Recognition}, pages 5414--5423, 2021.

\bibitem{esser2021taming}
Patrick Esser, Robin Rombach, and Björn Ommer.
\newblock Taming transformers for high-resolution image synthesis, 2021.

\bibitem{pascal}
Mark Everingham, Luc Van~Gool, Christopher~KI Williams, John Winn, and Andrew
  Zisserman.
\newblock The pascal visual object classes (voc) challenge.
\newblock {\em IJCV}, 88(2):303--338, 2010.

\bibitem{Caltech101}
Li Fei-Fei, Rob Fergus, and Pietro Perona.
\newblock Learning generative visual models from few training examples: An
  incremental bayesian approach tested on 101 object categories.
\newblock In {\em CVPR workshop}, pages 178--178. IEEE, 2004.

\bibitem{feng2020kd3a}
Hao-Zhe Feng, Zhaoyang You, Minghao Chen, Tianye Zhang, Minfeng Zhu, Fei Wu,
  Chao Wu, and Wei Chen.
\newblock Kd3a: Unsupervised multi-source decentralized domain adaptation via
  knowledge distillation.
\newblock {\em arXiv preprint arXiv:2011.09757}, 2020.

\bibitem{fer2013}
Ian~J Goodfellow, Dumitru Erhan, Pierre~Luc Carrier, Aaron Courville, Mehdi
  Mirza, Ben Hamner, Will Cukierski, Yichuan Tang, David Thaler, Dong-Hyun Lee,
  et~al.
\newblock Challenges in representation learning: A report on three machine
  learning contests.
\newblock In {\em International conference on neural information processing},
  pages 117--124. Springer, 2013.

\bibitem{guo2020dmcp}
Shaopeng Guo, Yujie Wang, Quanquan Li, and Junjie Yan.
\newblock Dmcp: Differentiable markov channel pruning for neural networks.
\newblock In {\em Proceedings of the IEEE/CVF Conference on Computer Vision and
  Pattern Recognition}, pages 1539--1547, 2020.

\bibitem{guo2020adafilter}
Yunhui Guo, Yandong Li, Liqiang Wang, and Tajana Rosing.
\newblock Adafilter: Adaptive filter fine-tuning for deep transfer learning.
\newblock In {\em Proceedings of the AAAI Conference on Artificial
  Intelligence}, volume~34, pages 4060--4066, 2020.

\bibitem{guo2019spottune}
Yunhui Guo, Honghui Shi, Abhishek Kumar, Kristen Grauman, Tajana Rosing, and
  Rogerio Feris.
\newblock Spottune: transfer learning through adaptive fine-tuning.
\newblock In {\em Proceedings of the IEEE/CVF Conference on Computer Vision and
  Pattern Recognition}, pages 4805--4814, 2019.

\bibitem{he2020momentum}
Kaiming He, Haoqi Fan, Yuxin Wu, Saining Xie, and Ross Girshick.
\newblock Momentum contrast for unsupervised visual representation learning.
\newblock In {\em Proceedings of the IEEE/CVF Conference on Computer Vision and
  Pattern Recognition}, pages 9729--9738, 2020.

\bibitem{he2017mask}
Kaiming He, Georgia Gkioxari, Piotr Doll{\'a}r, and Ross Girshick.
\newblock Mask r-cnn.
\newblock In {\em Proceedings of the IEEE international conference on computer
  vision}, pages 2961--2969, 2017.

\bibitem{he2016deep}
Kaiming He, Xiangyu Zhang, Shaoqing Ren, and Jian Sun.
\newblock Deep residual learning for image recognition.
\newblock In {\em Proceedings of the IEEE conference on computer vision and
  pattern recognition}, pages 770--778, 2016.

\bibitem{eurosat}
Patrick Helber, Benjamin Bischke, Andreas Dengel, and Damian Borth.
\newblock Eurosat: A novel dataset and deep learning benchmark for land use and
  land cover classification.
\newblock {\em IEEE Journal of Selected Topics in Applied Earth Observations
  and Remote Sensing}, 12(7):2217--2226, 2019.

\bibitem{hinton2015distilling}
Geoffrey Hinton, Oriol Vinyals, and Jeff Dean.
\newblock Distilling the knowledge in a neural network.
\newblock {\em arXiv preprint arXiv:1503.02531}, 2015.

\bibitem{houlsby2019parameter}
Neil Houlsby, Andrei Giurgiu, Stanislaw Jastrzebski, Bruna Morrone, Quentin
  De~Laroussilhe, Andrea Gesmundo, Mona Attariyan, and Sylvain Gelly.
\newblock Parameter-efficient transfer learning for nlp.
\newblock In {\em International Conference on Machine Learning}, pages
  2790--2799. PMLR, 2019.

\bibitem{howard2018universal}
Jeremy Howard and Sebastian Ruder.
\newblock Universal language model fine-tuning for text classification.
\newblock {\em arXiv preprint arXiv:1801.06146}, 2018.

\bibitem{huo2021wenlan}
Yuqi Huo, Manli Zhang, Guangzhen Liu, Haoyu Lu, Yizhao Gao, Guoxing Yang,
  Jingyuan Wen, Heng Zhang, Baogui Xu, Weihao Zheng, Zongzheng Xi, Yueqian
  Yang, Anwen Hu, Jinming Zhao, Ruichen Li, Yida Zhao, Liang Zhang, Yuqing
  Song, Xin Hong, Wanqing Cui, Danyang Hou, Yingyan Li, Junyi Li, Peiyu Liu,
  Zheng Gong, Chuhao Jin, Yuchong Sun, Shizhe Chen, Zhiwu Lu, Zhicheng Dou, Qin
  Jin, Yanyan Lan, Wayne~Xin Zhao, Ruihua Song, and Ji-Rong Wen.
\newblock Wenlan: Bridging vision and language by large-scale multi-modal
  pre-training, 2021.

\bibitem{kirkpatrick2017overcoming}
James Kirkpatrick, Razvan Pascanu, Neil Rabinowitz, Joel Veness, Guillaume
  Desjardins, Andrei~A Rusu, Kieran Milan, John Quan, Tiago Ramalho, Agnieszka
  Grabska-Barwinska, et~al.
\newblock Overcoming catastrophic forgetting in neural networks.
\newblock {\em Proceedings of the national academy of sciences},
  114(13):3521--3526, 2017.

\bibitem{kolesnikov2020big}
Alexander Kolesnikov, Lucas Beyer, Xiaohua Zhai, Joan Puigcerver, Jessica Yung,
  Sylvain Gelly, and Neil Houlsby.
\newblock Big transfer (bit): General visual representation learning.
\newblock In {\em Computer Vision--ECCV 2020: 16th European Conference,
  Glasgow, UK, August 23--28, 2020, Proceedings, Part V 16}, pages 491--507.
  Springer, 2020.

\bibitem{kou2020stochastic}
Zhi Kou, Kaichao You, Mingsheng Long, and Jianmin Wang.
\newblock Stochastic normalization.
\newblock {\em Advances in Neural Information Processing Systems}, 33, 2020.

\bibitem{stanfordcar}
Jonathan Krause, Michael Stark, Jia Deng, and Li Fei-Fei.
\newblock 3d object representations for fine-grained categorization.
\newblock In {\em 4th International IEEE Workshop on 3D Representation and
  Recognition (3dRR-13)}, Sydney, Australia, 2013.

\bibitem{cifar}
Alex Krizhevsky, Geoffrey Hinton, et~al.
\newblock Learning multiple layers of features from tiny images.
\newblock 2009.

\bibitem{kundu2020universal}
Jogendra~Nath Kundu, Naveen Venkat, R~Venkatesh Babu, et~al.
\newblock Universal source-free domain adaptation.
\newblock In {\em Proceedings of the IEEE/CVF Conference on Computer Vision and
  Pattern Recognition}, pages 4544--4553, 2020.

\bibitem{li2019delta}
Xingjian Li, Haoyi Xiong, Hanchao Wang, Yuxuan Rao, Liping Liu, Zeyu Chen, and
  Jun Huan.
\newblock Delta: Deep learning transfer using feature map with attention for
  convolutional networks.
\newblock {\em arXiv preprint arXiv:1901.09229}, 2019.

\bibitem{liang2020we}
Jian Liang, Dapeng Hu, and Jiashi Feng.
\newblock Do we really need to access the source data? source hypothesis
  transfer for unsupervised domain adaptation.
\newblock In {\em International Conference on Machine Learning}, pages
  6028--6039. PMLR, 2020.

\bibitem{COCO}
Tsung-Yi Lin, Michael Maire, Serge Belongie, James Hays, Pietro Perona, Deva
  Ramanan, Piotr Doll{\'a}r, and C~Lawrence Zitnick.
\newblock Microsoft coco: Common objects in context.
\newblock In {\em ECCV}, pages 740--755. Springer, 2014.

\bibitem{liu2021gpt}
Xiao Liu, Yanan Zheng, Zhengxiao Du, Ming Ding, Yujie Qian, Zhilin Yang, and
  Jie Tang.
\newblock Gpt understands, too.
\newblock {\em arXiv preprint arXiv:2103.10385}, 2021.

\bibitem{liu2021swin}
Ze Liu, Yutong Lin, Yue Cao, Han Hu, Yixuan Wei, Zheng Zhang, Stephen Lin, and
  Baining Guo.
\newblock Swin transformer: Hierarchical vision transformer using shifted
  windows.
\newblock {\em arXiv preprint arXiv:2103.14030}, 2021.

\bibitem{long2015learning}
Mingsheng Long, Yue Cao, Jianmin Wang, and Michael Jordan.
\newblock Learning transferable features with deep adaptation networks.
\newblock In {\em International conference on machine learning}, pages 97--105.
  PMLR, 2015.

\bibitem{loshchilov2017decoupled}
Ilya Loshchilov and Frank Hutter.
\newblock Decoupled weight decay regularization.
\newblock {\em arXiv preprint arXiv:1711.05101}, 2017.

\bibitem{mahajan2018exploring}
Dhruv Mahajan, Ross Girshick, Vignesh Ramanathan, Kaiming He, Manohar Paluri,
  Yixuan Li, Ashwin Bharambe, and Laurens Van Der~Maaten.
\newblock Exploring the limits of weakly supervised pretraining.
\newblock In {\em Proceedings of the European conference on computer vision
  (ECCV)}, pages 181--196, 2018.

\bibitem{aircraft}
S. Maji, J. Kannala, E. Rahtu, M. Blaschko, and A. Vedaldi.
\newblock Fine-grained visual classification of aircraft.
\newblock Technical report, 2013.

\bibitem{svhn}
Yuval Netzer, Tao Wang, Adam Coates, Alessandro Bissacco, Bo Wu, and Andrew~Y
  Ng.
\newblock Reading digits in natural images with unsupervised feature learning.
\newblock 2011.

\bibitem{flowers}
M-E Nilsback and Andrew Zisserman.
\newblock A visual vocabulary for flower classification.
\newblock In {\em CVPR}, volume~2, pages 1447--1454. IEEE, 2006.

\bibitem{pets}
Omkar~M Parkhi, Andrea Vedaldi, Andrew Zisserman, and CV Jawahar.
\newblock Cats and dogs.
\newblock In {\em CVPR}, pages 3498--3505. IEEE, 2012.

\bibitem{peng2019correlation}
Baoyun Peng, Xiao Jin, Jiaheng Liu, Shunfeng Zhou, Yichao Wu, Yu Liu, Dongsheng
  Li, and Zhaoning Zhang.
\newblock Correlation congruence for knowledge distillation, 2019.

\bibitem{radford2021learning}
Alec Radford, Jong~Wook Kim, Chris Hallacy, Aditya Ramesh, Gabriel Goh,
  Sandhini Agarwal, Girish Sastry, Amanda Askell, Pamela Mishkin, Jack Clark,
  et~al.
\newblock Learning transferable visual models from natural language
  supervision.
\newblock {\em arXiv preprint arXiv:2103.00020}, 2021.

\bibitem{rajasegaran2020self}
Jathushan {Rajasegaran}, Salman~H. {Khan}, Munawar {Hayat}, Fahad~Shahbaz
  {Khan}, and Mubarak {Shah}.
\newblock Self-supervised knowledge distillation for few-shot learning.
\newblock {\em arXiv preprint arXiv:2006.09785}, 2020.

\bibitem{ramachandran2016unsupervised}
Prajit Ramachandran, Peter~J Liu, and Quoc~V Le.
\newblock Unsupervised pretraining for sequence to sequence learning.
\newblock {\em arXiv preprint arXiv:1611.02683}, 2016.

\bibitem{ro2021autolr}
Youngmin Ro and Jin~Young Choi.
\newblock Autolr: Layer-wise pruning and auto-tuning of learning rates in
  fine-tuning of deep networks.
\newblock In {\em Proceedings of the AAAI Conference on Artificial
  Intelligence}, volume~35, pages 2486--2494, 2021.

\bibitem{sohn2020fixmatch}
Kihyuk Sohn, David Berthelot, Chun-Liang Li, Zizhao Zhang, Nicholas Carlini,
  Ekin~D Cubuk, Alex Kurakin, Han Zhang, and Colin Raffel.
\newblock Fixmatch: Simplifying semi-supervised learning with consistency and
  confidence.
\newblock {\em arXiv preprint arXiv:2001.07685}, 2020.

\bibitem{ucf101}
Khurram Soomro, Amir~Roshan Zamir, and Mubarak Shah.
\newblock Ucf101: A dataset of 101 human actions classes from videos in the
  wild.
\newblock {\em arXiv preprint arXiv:1212.0402}, 2012.

\bibitem{GTSRB}
Johannes Stallkamp, Marc Schlipsing, Jan Salmen, and Christian Igel.
\newblock The german traffic sign recognition benchmark: a multi-class
  classification competition.
\newblock In {\em The 2011 international joint conference on neural networks},
  pages 1453--1460. IEEE, 2011.

\bibitem{tian2019contrastive}
Yonglong Tian, Dilip Krishnan, and Phillip Isola.
\newblock Contrastive representation distillation.
\newblock {\em arXiv preprint arXiv:1910.10699}, 2019.

\bibitem{tian2020contrastive}
Yonglong Tian, Dilip Krishnan, and Phillip Isola.
\newblock Contrastive representation distillation, 2020.

\bibitem{vaswani2017attention}
Ashish Vaswani, Noam Shazeer, Niki Parmar, Jakob Uszkoreit, Llion Jones,
  Aidan~N Gomez, {\L}ukasz Kaiser, and Illia Polosukhin.
\newblock Attention is all you need.
\newblock In {\em Advances in neural information processing systems}, pages
  5998--6008, 2017.

\bibitem{sun}
Jianxiong Xiao, Krista~A Ehinger, James Hays, Antonio Torralba, and Aude Oliva.
\newblock Sun database: Exploring a large collection of scene categories.
\newblock {\em IJCV}, 119(1):3--22, 2016.

\bibitem{Yin_2020_CVPR}
Hongxu Yin, Pavlo Molchanov, Jose~M. Alvarez, Zhizhong Li, Arun Mallya, Derek
  Hoiem, Niraj~K. Jha, and Jan Kautz.
\newblock Dreaming to distill: Data-free knowledge transfer via deepinversion.
\newblock In {\em Proceedings of the IEEE/CVF Conference on Computer Vision and
  Pattern Recognition (CVPR)}, June 2020.

\bibitem{zhai2019large}
Xiaohua Zhai, Joan Puigcerver, Alexander Kolesnikov, Pierre Ruyssen, Carlos
  Riquelme, Mario Lucic, Josip Djolonga, Andre~Susano Pinto, Maxim Neumann,
  Alexey Dosovitskiy, et~al.
\newblock A large-scale study of representation learning with the visual task
  adaptation benchmark.
\newblock {\em arXiv preprint arXiv:1910.04867}, 2019.

\bibitem{zhang2018importance}
Jing Zhang, Zewei Ding, Wanqing Li, and Philip Ogunbona.
\newblock Importance weighted adversarial nets for partial domain adaptation.
\newblock In {\em Proceedings of the IEEE conference on computer vision and
  pattern recognition}, pages 8156--8164, 2018.

\bibitem{zhong2020regularizing}
Yang Zhong and Atsuto Maki.
\newblock Regularizing cnn transfer learning with randomised regression.
\newblock In {\em Proceedings of the IEEE/CVF Conference on Computer Vision and
  Pattern Recognition}, pages 13637--13646, 2020.

\end{thebibliography}
}

\end{document}